\documentclass{article}

\usepackage{arxiv}

\usepackage[utf8]{inputenc} 
\usepackage[T1]{fontenc}    
\usepackage{hyperref}       
\usepackage{url}            
\usepackage{booktabs}       
\usepackage{amsfonts}       
\usepackage{nicefrac}       
\usepackage{microtype}      
\usepackage{lipsum}
\usepackage{graphicx}
\graphicspath{ {./images/} }

\usepackage{amsmath}
\usepackage{amssymb}
\usepackage{mathtools}
\usepackage{amsthm}
\usepackage{tabularx}
\usepackage{rotating}
\usepackage{algorithm}
\usepackage{algorithmic}

\title{MEC-IP: Efficient Discovery of Markov Equivalent Classes \\ via Integer Programming}

\author{
 Abdelmonem Elrefaey \\
  School of Coumputing and Augmented Intelligence\\
  Arizona State University\\
  Tempe, AZ 85281 \\
  \texttt{aelrefae@asu.edu} \\
   \And
 Rong Pan \\
  School of Coumputing and Augmented Intelligence\\
  University of Pittsburgh\\
  Arizona State University\\
  Tempe, AZ 85281 \\
  \texttt{rpan1@asu.edu} \\
}

\begin{document}
\maketitle
\begin{abstract}
This paper presents a novel Integer Programming (IP) approach for discovering the Markov Equivalent Class (MEC) of Bayesian Networks (BNs) through observational data. The MEC-IP algorithm utilizes a unique clique-focusing strategy and Extended Maximal Spanning Graphs (EMSG) to streamline the search for MEC, thus overcoming the computational limitations inherent in other existing algorithms. Our numerical results show that not only a remarkable reduction in computational time is achieved by our algorithm but also an improvement in causal discovery accuracy is seen across diverse datasets. These findings underscore this new algorithm's potential as a powerful tool for researchers and practitioners in causal discovery and BNSL, offering a significant leap forward toward the efficient and accurate analysis of complex data structures.
\end{abstract}


\section{Introduction}
Most of the current machine learning and artificial intelligence tools are of a black-box nature and they primarily focus on prediction optimization, while falling short of model transparency and interpretability. Such properties, however, are often required by mission-critical applications such as in the areas of healthcare and governmental policy-making. In {\em The Book of Why}, the authors \cite{pearl2018book} repeatedly emphasize the importance of models being able to operate under causal frameworks to offer more than mere predictive insights. They propose a 'ladder of causation' with three tiers

\begin{itemize}
  \setlength\itemsep{0em}
    \item Level 1: Models that can only make predictive associations, such as forecasting symptoms for a specific disease.
    \item Level 2: Models that incorporate some level of causal understanding, allowing them to answer intervention-based questions.
    \item Level 3: Models that provide comprehensive causal insights, even extending to counterfactual reasoning.
\end{itemize}

These levels are also referred to as `seeing,' `doing,' and `imagining.' Bayesian Networks (BN), a concept Judea Pearl introduced decades earlier\cite{pearl1985bayesian,pearl2022reverend}, are equipped to tackle these questions at all levels, provided that they are used within a causal framework, i.e., causal BNs. These networks are probabilistic graphical models ideal for modeling complex, non-deterministic systems, and they have found applications across various fields from biology and healthcare to engineering and environmental science. A parametric BN is a graph that depicts direct (causal) dependencies between variables, where parameters in the BN define the form and strength of these relationships. It is known that specifying these parameters is generally simpler than accurately recovering the underlying graph, while learning the BN structure, also called `causal discovery', often requires both an in-depth knowledge of the system under study and a tremendous amount of observations \cite{kitson2023survey}. The problem of BN structure learning (BNSL) is considered the most challenging problem in parametric BN methods. This is because searching for the true graph is indeed an NP-Hard problem where some instances are much harder than others \cite{chickering2004large}.

There exists a plethora of BN strucutre learning algorithms. These algorithms range from early yet still effective methods to the latest breakthroughs, and they typically operate by searching over a set of possible graphs in some manner. Classical BNSL algorithms can be generally categorized into two classes. Firstly, the score-based methods represent a traditional machine learning approach where graphs are explored and scored in terms of how well the fitting distributions agree with the empirical distributions. The graph that maximizes the scoring function is returned as the preferred graph \cite{kitson2023survey}. On the other hand, the constraint-based learning approach is built upon a series of conditional independence tests that determine the addition, removal, or orientation of BN edges \cite{scutari2013identifying}. Additionally, there are hybrid algorithms being developed that adopt features from both score-based and constraint-based learning \cite{kitson2023survey}. Automating the construction of causal structures possesses a huge benefit to every research field that is concerned with causal inference and causal intervention \cite{nogueira2022methods}. However, the automated causal discovery is hindered by many difficulties that go beyond the problem of NP-hardness which can be generally addressed by pruning the search space of possible graphs and effectively minimizing the loss in accuracy and maximizing the gain in speed.

Learning the Directed Acyclic Graph (DAG) of a BN from data is only possible under certain assumptions \cite{koller2009probabilistic}. The first assumption is the \textit{Faithfulness Assumption}. This assumption states that all and only the conditional independencies that are present in data are implied by the true causal structure. In other words, if the data indicates that two variables are independent given a set of other variables, this independence should correspond to the true non-causal relationship, and vice versa. 
The second assumption, \textit{Causal Sufficiency}, is the assumption that is commonly used for facilitating the data-driven learning of BN structure. This assumption holds that all common causes of every variable included in the model are also included in the model. In other words, there are no hidden or unmeasured confounding variables that influence the variables being studied. This assumption is crucial because unmeasured confounders can lead to spurious associations, making it difficult to correctly infer causal relationships.
However, in real-world scenarios, especially in social sciences and biology, it is often challenging to ensure causal sufficiency due to the complexity of the systems and practical limitations in data collection \cite{10.1093/oxfordhb/9780199279739.003.0037}.

\subsection{The Proposed MEC-IP Algorithm}

 From an observational data study, it is often impossible to identify a unique DAG; instead, a class of DAGs can be equally good at fitting the available data. Markov Equivalent Class (MEC) is defined as the causal structure that encodes the same conditional independence relationships among variables, which, as the result of data fitting, is a Completed Partially Directed Acyclic Graph (CPDAG).  This concept is vital because, in many cases, the data alone cannot specify a unique DAG unless some other assumptions were made. Therefore, the goal of our algorithm in this paper is to identify the MEC.   

The exact scoring-based methods for BNSL are often computationally intensive, particularly as the number of variables increases. This is because they examine a large combinatorial space of potential network structures to find the best fit of data. Therefore, this difficulty is conventionally countered by setting a limit on the in-degree of each node in the graph, which results in suboptimal BNs. The MEC-IP algorithm, in contrast, adopts a data-driven targeted-clique-focusing approach. It employs Extended Maximal Spanning Graphs (EMSG) and iteratively refines the network while avoiding the exhaustive search inherent in the exact scoring-based methods. This efficient learning process is especially beneficial for larger datasets where computational resources and time are significant considerations. Moreover, while constraint-based methods, such as the PC algorithm, are known for their relative efficiency, they typically require a large number of samples and a large number of statistical tests to determine the conditional independence between every variable pair. This can be problematic in cases of limited data, as the reliability of these statistical tests diminishes with smaller sample sizes \cite{cai2022improving}. Moreover, extensive statistical tests can be time-consuming too. The algorithm we proposed, however, strategically selects which statistical tests to perform, focusing primarily on those most likely to yield informative results. By doing that, it not only reduces the computational burden but also mitigates the issue of unreliable results due to limited data. This data-driven approach to testing is particularly advantageous in scenarios where data is scarce or when the dataset is large but has a limited number of observations per variable.

The MEC-IP algorithm is a hybrid algorithm, which possesses notable advantages over both exact scoring-based methods, such as GOBNILP, and traditional constraint-based methods, such as the PC algorithm. These advantages are primarily in terms of efficiency and accuracy, especially when dealing with datasets of limited size. In summary, the MEC-IP algorithm offers a balanced approach that enhances both computational efficiency and outcome reliability. This makes it a valuable tool for causal discovery and BNSL.

In the remainder of the paper, the literature on existing BNSL algorithms is reviewed in Section \ref{Literature Review}. The details of the MEC-IP algorithm are explained in Section \ref{Methodology}. Section \ref{Num_results} presents experiments and numerical results. Lastly, Section \ref{conclusion} concludes the paper.

\section{Literature Review}
\label{Literature Review}

In the realm of BN structure learning, algorithms can categorized into either the constraint-based approach or the score-based approach, and finally the hybrid approach. 

\subsection{Constraint-based Approach}
The constraint-based algorithms are the cornerstones of BNSL. They utilize conditional independence tests to discern the graph that underpins causal or associative interactions among variables. Among the seminal works in this area are the SGS (Spirtes-Glymour-Scheines) \cite{spirtes1990causality} algorithm and the PC (Peter-Clark) \cite{spirtes1991algorithm} family of algorithms. Each of them contributes unique methodologies, strengths, and limitations to the field. The SGS algorithm is initiated with a completely connected undirected graph, and then it iteratively removes edges that fail conditional independence tests. 
It has found applications in diverse domains such as bioinformatics, particularly in identifying gene regulatory  \cite{scutari2013identifying}, and in economics for analyzing market dynamics \cite{book}. \cite{spirtes2014uniformly} also presents a conservative version of the SGS algorithm. 

On the other hand, the PC family of algorithms are built upon the SGS algorithm by incorporating heuristics to decrease computational complexity. Starting much like SGS with a fully connected undirected graph, the PC algorithm employs some more efficient edge-pruning procedures. Various adaptations exist within the PC family, including the Conservative PC (CPC) \cite{ramsey2012adjacency}, which adopts a more cautious approach in edge elimination, and the Fast Causal Inference (FCI) algorithm \cite{spirtes2001anytime}, which is equipped to handle latent variables and selection bias. However, despite these advancements, the PC family has its own set of challenges. Mainly, it is susceptible to the issues of false positives and false negatives, often stemming from data limitations or erroneous assumptions concerning conditional independence.
There exists other algorithm that belong to the PC family such as PC-stable \cite{colombo2014order}, and PC-MAX \cite{ramsey2016improving} that offer improvements related to consistency and accuracy of the results, respectively.


\subsection{Score-based Approach}
The score-based algorithms for BNSL offer a different approach from the constraint-based counterparts. Instead of relying on conditional independence tests to prune individual edges, the score-based algorithms evaluate the goodness-of-fit of an entire network to the data using scoring metrics such as the Bayesian Information Criterion (BIC) or the Akaike Information Criterion (AIC). This class of algorithms is often further classified into two major types: approximate methods, including Tabu Search and Hill Climbing (HC), as well as exact methods like Dynamic Programming (DP) \cite{yuan2011learning} and Integer Programming (IP). HC \cite{Bouckaert1994PropertiesOB} employs a greedy strategy, at each step choosing the neighboring solution that most improves the score, until no further improvement is possible. Tabu Search \cite{bouckaert1995bayesian} extends basic local search by maintaining a ``tabu list" of recently visited solutions to avoid cycling back, thus allowing for a more extensive search of the solution space. Tabu search and HC are often preferred for their computational efficiency, especially in high-dimensional scenarios. These methods, while fast and easy to implement, do not guarantee a global optimal solution and often settle in local optima. Their performance depends significantly on the choice of scoring function and network initiation. In contrast, exact methods like DP and GOBNILP (Globally Optimal Bayesian Network Learning with Integer Programming) aim to find the globally optimal structure. Dynamic programming \cite{koivisto2004exact,ott2003finding} techniques decompose the problem into overlapping sub-problems and recursively solve them, storing intermediate results for re-use. Although this approach guarantees finding the optimal solution, it is often too computationally expensive to be feasible for large networks. GOBNILP \cite{cussens2012bayesian,BARTLETT2017258,ijcai2017p708} reformulates the BNSL problem to be an Integer Linear Programming (ILP) problem. It uses mathematical optimization techniques to find the globally optimal solution, ensuring the quality of the learned network.

When comparing approximate and exact methods, one notices a trade-off between computational efficiency and solution quality. Approximate methods like HC are quicker but may yield suboptimal solutions, whereas GOBNILP is computationally intensive but guarantee global optimality. Thus, the choice between the two largely depends on the specific requirements of the application at hand.

\subsection{Hybrid Methods}
Hybrid algorithms for BNSL present a compelling middle-ground between constraint-based and score-based methodologies, effectively incorporating the strengths of both approaches to overcome individual limitations. Typically, these hybrid algorithms use constraint-based techniques for initial structure learning and then refine the structure using score-based methods. In this literature review, we focus on some well-known hybrid methods such as the Max-Min Hill Climbing (MMHC) algorithm proposed by Tsamardinos et al. \cite{tsamardinos2006max} and the RSMAX2 algorithm, among others. 

Max-Min Hill Climbing (MMHC) is a seminal hybrid method that first uses a constraint-based phase to learn an initial skeleton of the network. Then, it uses the max-min parents-and-children (MMPC) algorithm to perform skeleton tuning. After skeleton identification, the algorithm switches to a score-based phase, employing HC to optimize the network structure. This phased approach allows MMHC to benefit from the computational efficiency of constraint-based methods while also leveraging the optimization capabilities of score-based techniques. However, MMHC's performance can be affected by the quality of the initial skeleton and the scoring function used in the second phase. It has been widely applied in bioinformatics, particularly in learning gene networks \cite{song2022using}. Another noteworthy hybrid algorithm is RSMAX2, which combines restricted search space-based constraint techniques with Bayesian scoring mechanisms. It applies a score-based algorithm within restricted search spaces identified through constraint-based methods, effectively utilizing the advantages of both approaches. This algorithm often yields results comparable to exact methods while maintaining computational efficiency similar to approximate methods. Nevertheless, the performance of RSMAX2 depends on the accuracy of the restricted search spaces and may vary accordingly. Other emerging hybrid methods focus on the scheme of adaptively switching between constraint-based and score-based phases. These methods aim to dynamically capitalize on the strengths of each approach, although they often require intricate parameter tuning and a deep understanding of the problem domain. 

\section{The MEC-IP Algorithm}
\label{Methodology}
The MEC-IP algorithm employs a systematic approach to learning the BN through data, ensuring the learned structure accurately represents the underlying causal relationships. Initially, it performs a $\chi^2$ test for independence on all variable pairs to assess their direct interactions. Based on the test statistics, these pairs are then ranked, leading to the formation of an EMSG by selectively removing weaker edges according to their comparative $\chi^2$ values with shared neighbors. Subsequently, edges with a p-value above a significance threshold are eliminated, resulting in an Undirected Graph (UG).

From this UG, a CPDAG is formulated through an IP by orienting the edges. This CPDAG is thus representing the initial MEC of DAGs. Our algorithm will then iteratively refine the CPDAG. First, it identifies candidate triangular cliques and tests for conditional independence using $\chi^2$ tests, thereby determining whether additional connections are necessary to explain the dependency between node pairs. If so, new edges are added. This process is repeated, and the IP is resolved to refine the graph's structure. This iterative process continues until no further improvements can be made, culminating in a final CPDAG that best fits the data, representing the MEC of most suitable DAGs.

The MEC-IP algorithm consists of the following main steps:

\begin{enumerate}
    \item The $\chi^2$ test statistic for (unconditional) independence is calculated for every pair of variables (edge). 
    
    \item The edges are ordered in ascending order of the $\chi^2$ statistic. The EMSG is produced by removing edges between two nodes \( A \) and \( B \) if and only if \( A \) and \( B \) share a neighbor \( C \) where 
    \[ \chi^2(A \leftrightarrow B) < \chi^2(A \leftrightarrow C) \] and
    \[ \chi^2(A \leftrightarrow B) < \chi^2(B \leftrightarrow C) \]
    
    \item From the remaining edges, the edges with a \( p \)-value less than the required significance threshold are dropped. By now, an Undirected Graph (UG) is formed.

    \item Using the UG as input, the initial MEC of DAGs with the highest BIC score, represented by a CPDAG, is constructed by solving the BNSL IP.
        
    \item For every candidate triangular clique in the CPDAG, one of the minimum \( d \)-separating node sets is identified such that the pair of nodes in the triplet that are not currently joined are \( d \)-separated in the CPDAG.
    
    \item A $\chi^2$ conditional independence test between the pair given the \( d \)-separating node set is performed.
    
    \item If the test is found to reject the null hypothesis, the current graph is deemed insufficient in explaining the dependence relationships between the pair of nodes, and an undirected edge is added between them. Otherwise, they remain unconnected.
    
    \item After going through all candidate triplets, the IP is solved again to orient the added edges as a result of the triangulation phase, and a new CPDAG is returned.
    
    \item Repeat Steps 5-8 until no further improvement can be achieved. The final CPDAG, representing the MEC of the best data-fitting DAGs, is returned as the final output.

\end{enumerate}

\subsection{EMSG}

Notice that the initial EMSG is obtained from the unconditional test of independence between the variables. The ESMG generating algorithm is highly motivated by the work in \cite{constantinou2020learninga, constantinou2020learningb} and it is summarized in Algorithm \ref{alg:emsg}. This algorithm returns a UG that will be used in subsequent steps of the MEC-IP algorithm.

\begin{algorithm}[h]
   \caption{The EMSG Generation Algorithm}
   \label{alg:emsg}
\begin{algorithmic}
   \STATE {\bfseries Input:} score list $score$
   \STATE {\bfseries{Initialize graph $G$ as a completed graph.}}
   \STATE {\bfseries Initialize edge-weights:}
   \FOR{each edge-weight pair $(edge, weight)$ in $edges\_weights$}
   \STATE Add edge $edge$ with weight $weight$ to graph $G$.
   \ENDFOR
   \STATE {\bfseries Edge removal:}
   \FOR{each edge $(A, B)$ in $edges\_weights$}
   \IF{edge $(A, B)$ exists in $G$}
   \FOR{each common neighbor $C$ of $A$ and $B$ in $G$}
   \IF{$(weight(A,C) > weight(A,B) < weight(B,C))$ and edge $(A,C)$ exists and edge $(B,C)$ exists}
   \STATE Remove edge $(A, B)$ from $G$.
   \STATE Break.
   \ENDIF
   \ENDFOR
   \ENDIF
   \ENDFOR
   \STATE {\bfseries Output:} List of remaining undirected edges in $G$.
\end{algorithmic}
\end{algorithm}

\subsection{Bayesian Network Structure Learning IP Model}

The IP steps in the MEC-IP algorithm \ref{alg:bn_structure_ip} specifically incorporate variables that adhere to the UG formed from the  EMSG step. This focus ensures that only significant and relevant relationships between variables, as determined in EMSG, are considered in the BNSL process.

To illustrate how this works, consider a simple example of a 4-node network consisting of nodes A, B, C, and D. Assume that the EMSG step has resulted in a UG with the following edges:

\begin{itemize}
  \setlength\itemsep{0em}
\centering
    \item $A \leftrightarrow B$
    \item $B \leftrightarrow C$
    \item $C \leftrightarrow D$
\end{itemize}

In this scenario, the EMSG has identified that nodes A and B, B and C, and C and D have significant correlations. However, there's no direct significant correlation between A and C, A and D, or B and D. In the IP model, the algorithm will consider parent sets for each node that are consistent with this UG structure. For instance, for node A, its potential parents could be B (since $A \leftrightarrow B$ exists in the UG).
For node B, its potential parents could be A or C or both. For node C, its potential parents could be B or D or both. For node D, its potential parents could be C. In addition, each node could also have an empty parent set (i.e. this node is a root node in the graph). This targeted calculation is in contrast with a scenario where one would compute the BIC local scores for every possible variable-parent combination in the network, which, in the case of a 4-node network, would include additional pairs such as A-C, A-D, and B-D. By focusing only on the relationships established by the EMSG, the algorithm reduces the computational burden.

The IP model then works to optimize the network structure under these constraints, seeking the configuration that best fits the data while adhering to the relationships already specified by the EMSG. It does so by maximizing the objective function, which is the sum of the BIC scores reflecting how well each parent set explains the data. This approach ensures that the learned BN structure is not only data-driven but also respects the conditional independencies inferred during the EMSG stage. The aim is to select the best parent group for each variable. This selection is done in such a way as to maximize the overall explanatory power of the network, which is quantified by the sum of the scores of the chosen parent groups. Note, unlike other IP algorithms, our algorithm does not limit the number of parent nodes. 
 
\begin{algorithm}[htb]
\caption{BN Structure Learning via IP}
\label{alg:bn_structure_ip}
\begin{algorithmic}
\STATE {\bfseries Input:} Local scores for potential parent sets, data table with variables.

\STATE {\bfseries Initialize Model and Variables:}
\STATE\hspace{12pt} Create an IP model for BN learning.
\STATE\hspace{12pt} Define variables as nodes in the BN, derived from data columns and UG from EMSG.
\STATE\hspace{12pt} Determine local scores for each node based on potential parent sets.

\STATE {\bfseries Define Model Constraints:}
\STATE\hspace{12pt} Ensure each node has exactly one parent set through constraints.
\STATE\hspace{12pt} Optionally, add constraints to prevent cycles in the network.

\STATE {\bfseries Set Objective Function:}
\STATE\hspace{12pt} Define the objective function as the maximization of the sum of local scores for selected parent sets.

\STATE {\bfseries Iterative Model Solving and Cycle Checking:}
\WHILE{true}
\STATE Solve the IP model.
\STATE Construct a DAG from the current solution.
\STATE Check the DAG for cycles.
\IF{cycles are found}
\STATE Formulate and add constraints to the model to break these cycles.
\ELSE
\STATE Exit the loop.
\ENDIF
\ENDWHILE
\STATE {\bfseries Output:} The optimized BN structure and an objective value (BIC score) of the model.
\end{algorithmic}
\end{algorithm}

The detailed IP is as follows:

\subsection*{Objective Function:}
Maximize:
\begin{equation*}
\sum_{(node, S) \in \text{node-parent set pairs}} \text{score}_{node, S} \cdot x_{node, S}
\end{equation*}
where \( x_{node, S} \) is a binary variable indicating whether parent set \( S \) is chosen for node \( node \).

The objective function in this BNSL method is deeply intertwined with the concept of \emph{decomposable} scores, which means that the total score of a network can be decomposed to a set of individual scores for each variable and its corresponding parent set. Such an additive nature allows for efficient optimization because changes in the parent set of one variable do not directly affect the scores of others. The BIC score is an example of a score that is decomposable. 

\subsection*{Constraints:}

There is a crucial condition that the resulting network must adhere to -- the resulting network cannot have any cycles. A cycle in a network would imply that a variable can, directly or indirectly, be its own ancestor, which is not permissible in a BN. To ensure this condition, a \emph{cutting plane} regime is used iteratively. The IP is repeatedly solved, each time checking if the resulting network has any cycles using the algorithm given by \cite{johnson1975finding}. If a cycle is found, we modify the mathematical model by adding the corresponding constraint (cutting plane) that prevents this cycle and then solve the IP again. This process continues until it finds a solution where the network has no cycles.

\begin{itemize}
    \item \textbf{One Parent Set Per Variable:}
    \begin{equation*}
    \sum_{S \in \text{parent sets of node}} x_{node, S} = 1 \quad \forall \text{ node}
    \end{equation*}

    \item \textbf{Cycle Prevention:} (iteratively added as needed)
    \begin{equation*}
    \sum_{i \in \{c_l\}_l} \sum_{S \in \text{parent sets of } i \text{ with } S \cap \{c_l\}_l = \emptyset} x_{i, S} \geq 1
    \end{equation*}
    for each detected cycle \( \{c_l\}_l \).
\end{itemize}

\subsection*{Decision Variables:}
\begin{equation*}
x_{node, S} \in \{0, 1\}
\end{equation*}
for each node and its corresponding parent sets.

In the end, a BN is constructed that both maximizes the explanatory scores and adheres to the acyclic structure requirement. This network represents the most likely relationships between variables, as dictated by the data and the initial scores.

Note that the IP returns a DAG representation of the BN learned and that the BIC score is an example of an \emph{equivalent} score. In the context of model selection, score equivalence means that different models (e.g., different structures of a BN) that encode the same set of conditional independencies have the same score. Correspondingly, the score of the returned DAG is one possible DAG within the MEC that represent those set of conditional indepencies. As such, any other DAG that belongs to the same MEC will have the same BIC score. Therefore, these DAGs are transformed to the CPDAG that represents the MEC. 

To transform a DAG of a BN into its equivalent CPDAG, a systematic approach is employed that emphasizes preserving the conditional independence relationships inherent in the original DAG. This process begins by identifying all instances of \emph{immorality} within the DAG—situations where two nodes share a common child but have no direct link themselves. These structures are pivotal because altering them would change the graph's conditional independencies. Then, the skeleton of the DAG is constructed by removing the directionality from all edges, thereby focusing on the structure without the causal directions. The next step involves reorienting edges in this skeleton to reflect the identified immoralities, ensuring these critical configurations are maintained in the CPDAG. Further refinement of edge orientations is achieved through the application of Meek's Rules. These rules systematically infer additional edge directions based on the principle of avoiding the creation of cycles or new immoralities, leveraging the current directed and undirected edges' configuration. The rules are applied iteratively to direct as many of the remaining undirected edges as possible without contradicting the original DAG's conditional independence properties. This iterative application of Meek's Rules helps orient edges in such a way that the resulting CPDAG accurately represents the MEC of the original DAG. The CPDAG thus captures all DAGs that are observationally indistinguishable from the original, showcasing the same conditional independencies among the variables. This process ensures the CPDAG serves as a comprehensive representation, encapsulating the range of potential causal structures compatible with the observed data, all while adhering to the conditional independencies defined by the initial DAG.

The CPDAG is then used to derive additional conditional independence relationships between the variables in the subsequent step of the MEC-IP algorithm, called \emph{triangulation}.  

\subsection{Triangulation}

The triangulation step plays a crucial role in enriching the network's complexity by methodically adding undirected edges. This phase addresses the initial network's potential under-specification problem as a result of employing the EMSG algorithm, ensuring that significant dependencies between nodes are not overlooked. During triangulation, the algorithm identifies potential undirected edges that could complete triangular cliques in the current network. The identification of these edges is based on a thorough examination of the network's existing structure, focusing on how adding an edge might reveal previously unrepresented relationships.

During the triangulation step of the MEC-IP algorithm, the concept of \emph{d-separation} \cite{pearl1988probabilistic} is essential to determining the conditional independence of nodes. The d-separation in BN refers to the criterion used to assess whether a set of nodes is independent of another set, given a third set (the conditioning set). This assessment will add an undirected edge to the network if two nodes are not d-separated. Specifically, for a potential edge between two nodes (say A and B), the algorithm examines all paths in the network between these nodes. Each path is assessed to see if it is blocked or unblocked based on the principle of d-separation:

\begin{itemize}
  \setlength\itemsep{0em}
    \item A path is considered blocked if it contains a node where the incoming and outgoing arrows meet head-to-tail or tail-to-tail, and this node is included in the conditioning set.
    \item Conversely, a path is also blocked if it contains a node where arrows meet head-to-head, and neither the node nor any of its descendants are in the conditioning set.
\end{itemize}

If the $\chi^2$ conditional independence test finds that nodes A and B are independent given the d-separating set, it implies that the potential edge between them is unnecessary; otherwise, the edge should be added. After thickening the network with potential edges, the IP model is solved again to determine the optimal directions of these new edges, ensuring that the learned network accurately represents the directional dependencies implied by the data. This iterative process of edge addition followed by IP resolution ensures that the final network structure is both comprehensive and representative of the underlying causal relationships.

\section{Numerical Experiments}
\label{Num_results}
We conducted experiments on multiple well-known BNs \cite{scutari2009learning, scutari2012bayesian}. The parameters of these BNs can be found in Table \ref{bnlearn}.

\begin{table}[h]
\caption{Test Network Parameters.}
\label{bnlearn}
\vskip 0.15in
\begin{tabularx}{\columnwidth}{p{1cm}p{.5cm}p{.5cm}p{2cm}p{3cm}p{12cm}}
\toprule
\textbf{Network} & \textbf{Nodes} & \textbf{Arcs} & \textbf{Avg Degree} & \textbf{Max In-Degree} & \textbf{Application} \\ \hline
Asia & 8 & 8 & 2.00 & 2 & Educational purposes, small-scale health studies \\
Sachs & 11 & 17 & 3.09 & 3 & Bioinformatics, cellular signaling pathways \\
Child & 20 & 25 & 2.50 & 2 & Modeling child health, diseases and symptoms \\
Insurance & 27 & 52 & 3.85 & 3 & Risk assessment in the insurance industry \\
Water & 32 & 66 & 4.12 & 5 & Water resource management \\
ALARM & 37 & 46 & 2.49 & 4 & Patient monitoring and diagnosis in medical field \\
Barley & 48 & 84 & 3.50 & 4 & Agricultural research, factors affecting barley yield \\
Win95pts & 76 & 112 & 2.95 & 7 & Analyzing failure points in the Windows 95 OS \\
\bottomrule
\end{tabularx}
\vskip -0.1in
\end{table}

In these experiments, we evaluate and compare four algorithms -- MEC-IP, Gobnilp \cite{pmlr-v138-cussens20a}, HC (hill climbing with Tabu list), and PC \cite{neapolitan2004learning} -- for learning BNs across various networks, where each of them has different characteristics and used in different applications. Three evaluation metrics are applied: Missing \% (the percentage of missing arcs), Extra \% (the percentage of extra arcs not present in the original network), and Total Time (the computing time, in seconds, taken by the algorithm to learn the network). Lower values in these metrics indicate better performance. Our study was conducted with different sample sizes (1000 and 10000 observations) for each network. Each trial was repeated 10 times with random datasets generated from the BN. Note that Gobnilp was run with a in-degree limit of 2 parents for each variable to avoid memory issues and excessive computational time.

The detailed results can be found in Table \ref{Network-table-detailed} in the Appendix. For small networks, such as Asia, Sachs, and Child, Gobnilp and MEC-IP often excel, demonstrating high accuracy with reasonably good computational efficiency. However, as the complexity and size of the networks increased like in the Insurance or Water BNs, MEC-IP showed a better balance between accuracy and computational time. Gobnilp, while maintaining high accuracy, tended to require significantly more computational resources, particularly in larger networks with larger sample sizes. Generally, the performance of all algorithms can be improved with larger sample sizes, manifesting in lower missing and extra percentages. This improvement, however, came at the cost of increased computational time, particularly notable in algorithms like MEC-IP and PC in larger networks. This trade-off between accuracy and computational efficiency was a recurring theme across the study. Among the algorithms, HC, despite being faster, often lagged in accuracy, exhibiting higher missing and extra percentages. This makes it less favorable for applications where precision is crucial. On the other hand, the PC algorithm often requires substantial computational time and does not provide a consistent performance in terms of its accuracy, as compared to other algorithms.

A second study, to which the results were summarized in Tables \ref{20 - Networks} and \ref{60 - Networks} in the Appendix, was also conducted to examine the performance of the four algorithms in learning a set of synthetic BNs with various structural properties. These BNs were randomly generated with their structural parameters specified by a tuple of four values, including the number of nodes (20 or 60), the maximum in-degree allowed (2 or 3), the maximum number of states that a variable can have (2 or 4), and the strength of the relationship between variables (1 for higher strength, 5 for lower strength). In this simulation study, each algorithm exhibited distinct performance characteristics. The MEC-IP algorithm stood out for its consistent and robust performance across various network settings, achieving overall lower Missing and Extra percentages, and particularly effective in networks with stronger inter-variable relationships. Gobnilp also performed commendably, particularly in smaller networks and those with strong relationships, though its performance quickly deteriorates in larger and more complex networks. The HC algorithm, while faster in some instances, tended to have higher error rates, making it suitable for rapid, approximate solutions but less ideal for applications that demand high precision. Conversely, the PC algorithm often showed higher Missing but lower Extra percentages, indicating a cautious approach to predicting new connections, thus favoring the scenarios where avoiding false connections is prioritized over comprehensively identifying all edges.

\begin{figure}[h]
\vskip 0.2in
\begin{center}
\centerline{\includegraphics[width=\columnwidth,height=250pt]{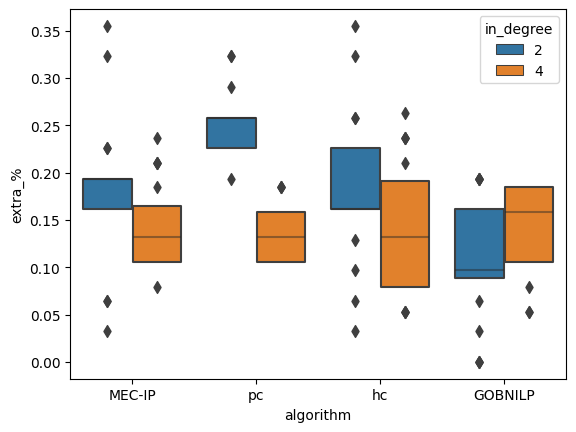}}
\caption{Extra\_\% Distribution Across Simulations by Network Complexity and Algorithm}
\label{extra-bp}
\end{center}
\vskip -0.2in
\end{figure}

Figures \ref{missing-bp} and \ref{extra-bp} analysis of how the complexity of Bayesian networks, as determined by the maximum in-degree (2 or 4), affects the performance of four structure learning algorithms. The Missing \% boxplot indicates that networks with an in-degree of 4 present a greater challenge for all algorithms, with GOBNILP showing a noticeable increase in the Missing \%, suggesting it find it particularly challenging to detect all true connections as complexity rises. Of course this is to be expected as the maximum in-degree was set to 2 for the purposes of this study. In contrast, MEC-IP, PC and HC display a smaller increase in Missing \% with higher in-degree, indicating a degree of robustness to complexity; however, the spread of results, especially for HC, points to variability in performance. The Extra \% boxplot shows MEC-IP has a decrease in false positives with the higher in-degree, while GOBNILP behaves in the opposite direction. In addition, MEC-IP's performance remains relatively stable across both levels of in-degree, indicating consistency in its approach to inferring connections regardless of network complexity. 


\begin{figure}[h]
\vskip 0.2in
\begin{center}
\centerline{\includegraphics[width=\columnwidth,height=250pt]{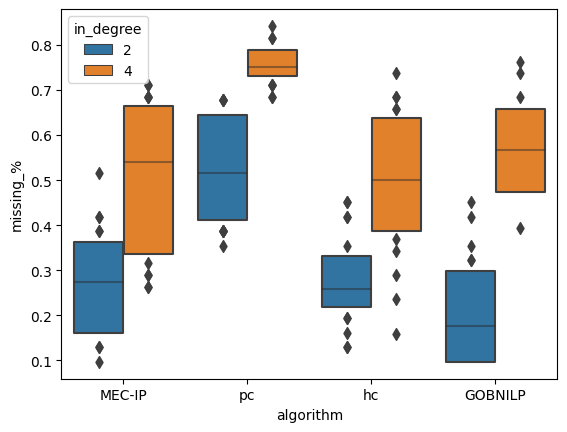}}
\caption{Missing\_\% Distribution Across Simulations by Network Complexity and Algorithm}
\label{missing-bp}
\end{center}
\vskip -0.2in
\end{figure}

Furthermore, figures \ref{missing-size} and \ref{extra-size} in the appendix illustrate the distribution of the missing and extra percentage of edges across the simulated networks. From these plots, we observe a general trend of increasing sample size leading to a decrease in both Extra \% and Missing \% for all algorithms, suggesting that having more data will enhance an algorithm's accuracy in inferring the true structure of the network by reducing false positives and false negatives. Specifically, MEC-IP shows a commendable consistency in its performance, with a narrow interquartile range (IQR) that indicates fewer variations between networks, and it benefits significantly from larger samples, as evidenced by the reduced median Extra \% and Missing \%. PC displays a greater variability. HC, on the other hand, seems to have a higher propensity for inferring more extra edges when the data is limited, demonstrated by a higher median and more outliers for the smaller sample size. GOBNILP stands out for its relatively stable performance across different sample sizes; its median Extra \% and Missing \% show minimal changes, suggesting that its performance is less sensitive to the data size compared to the other algorithms. This stability could be advantageous in situations where data is scarce.

\section{Conclusion}
\label{conclusion}
The MEC-IP algorithm is an improved IP approach to BNSL. It can efficiently identify MECs with reduced computation and fewer missing edge or extra edge errors. More importantly, it does not limit the number of parents of each node, thus suitable even for dense networks that have not been handled by previous IP approaches. Our numerical evaluation, comparing MEC-IP against several other established algorithms across multiple datasets, demonstrates its robustness and effectiveness, particularly in complex and large-scale networks. 

Future work will explore further optimizations and the applications of MEC-IP in real-world scenarios, enhancing its utility for causal inference and decision-making processes in various fields. This research contributes to the ongoing development of more efficient and accurate methods for causal discovery, fostering deeper insights into the underlying mechanisms of complex systems.

\bibliography{references.bib}

\begin{thebibliography}{10}

\bibitem{pearl2018book}
Judea Pearl and Dana Mackenzie.
\newblock {\em The book of why: the new science of cause and effect}.
\newblock Basic books, 2018.

\bibitem{pearl1985bayesian}
Judea Pearl.
\newblock Bayesian networks: A model of self-activated memory for evidential reasoning.
\newblock In {\em Proceedings of the 7th conference of the Cognitive Science Society, University of California, Irvine, CA, USA}, pages 15--17, 1985.

\bibitem{pearl2022reverend}
Judea Pearl.
\newblock Reverend bayes on inference engines: A distributed hierarchical approach.
\newblock In {\em Probabilistic and Causal Inference: The Works of Judea Pearl}, pages 129--138. 2022.

\bibitem{kitson2023survey}
Neville~Kenneth Kitson, Anthony~C Constantinou, Zhigao Guo, Yang Liu, and Kiattikun Chobtham.
\newblock A survey of bayesian network structure learning.
\newblock {\em Artificial Intelligence Review}, pages 1--94, 2023.

\bibitem{chickering2004large}
Max Chickering, David Heckerman, and Chris Meek.
\newblock Large-sample learning of bayesian networks is np-hard.
\newblock {\em Journal of Machine Learning Research}, 5:1287--1330, 2004.

\bibitem{scutari2013identifying}
Marco Scutari and Radhakrishnan Nagarajan.
\newblock On identifying significant edges in graphical models of molecular networks, 2013.

\bibitem{nogueira2022methods}
Ana~Rita Nogueira, Andrea Pugnana, Salvatore Ruggieri, Dino Pedreschi, and Jo{\~a}o Gama.
\newblock Methods and tools for causal discovery and causal inference.
\newblock {\em Wiley interdisciplinary reviews: data mining and knowledge discovery}, 12(2):e1449, 2022.

\bibitem{koller2009probabilistic}
Daphne Koller and Nir Friedman.
\newblock {\em Probabilistic graphical models: principles and techniques}.
\newblock MIT press, 2009.

\bibitem{10.1093/oxfordhb/9780199279739.003.0037}
Harold Kincaid.
\newblock {726 Causation in the Social Sciences}.
\newblock In {\em {The Oxford Handbook of Causation}}. Oxford University Press, 11 2009.

\bibitem{cai2022improving}
Erica Cai, Andrew McGregor, and David Jensen.
\newblock Improving the efficiency of the pc algorithm by using model-based conditional independence tests.
\newblock {\em arXiv preprint arXiv:2211.06536}, 2022.

\bibitem{spirtes1990causality}
P~Spirtes, C~Glymour, and R~Scheines.
\newblock causality from probability, proceedings of advanced computing for the social sciences.
\newblock {\em Williamsburgh, VA}, 1990.

\bibitem{spirtes1991algorithm}
Peter Spirtes and Clark Glymour.
\newblock An algorithm for fast recovery of sparse causal graphs.
\newblock {\em Social science computer review}, 9(1):62--72, 1991.

\bibitem{book}
Peter Spirtes, Clark Glymour, and Richard Scheines.
\newblock {\em Causation, Prediction, and Search}, volume~81.
\newblock 01 1993.

\bibitem{spirtes2014uniformly}
Peter Spirtes and Jiji Zhang.
\newblock A uniformly consistent estimator of causal effects under the k-triangle-faithfulness assumption.
\newblock {\em Statistical Science}, pages 662--678, 2014.

\bibitem{ramsey2012adjacency}
Joseph Ramsey, Jiji Zhang, and Peter~L Spirtes.
\newblock Adjacency-faithfulness and conservative causal inference.
\newblock {\em arXiv preprint arXiv:1206.6843}, 2012.

\bibitem{spirtes2001anytime}
Peter Spirtes.
\newblock An anytime algorithm for causal inference.
\newblock In {\em International Workshop on Artificial Intelligence and Statistics}, pages 278--285. PMLR, 2001.

\bibitem{colombo2014order}
Diego Colombo, Marloes~H Maathuis, et~al.
\newblock Order-independent constraint-based causal structure learning.
\newblock {\em J. Mach. Learn. Res.}, 15(1):3741--3782, 2014.

\bibitem{ramsey2016improving}
Joseph Ramsey.
\newblock Improving accuracy and scalability of the pc algorithm by maximizing p-value.
\newblock {\em arXiv preprint arXiv:1610.00378}, 2016.

\bibitem{yuan2011learning}
Changhe Yuan, Brandon Malone, and Xiaojian Wu.
\newblock Learning optimal bayesian networks using a* search.
\newblock In {\em Twenty-second international joint conference on artificial intelligence}, 2011.

\bibitem{Bouckaert1994PropertiesOB}
Remco~R. Bouckaert.
\newblock Properties of bayesian belief network learning algorithms.
\newblock In {\em Conference on Uncertainty in Artificial Intelligence}, 1994.

\bibitem{bouckaert1995bayesian}
Remco~Ronaldus Bouckaert.
\newblock {\em Bayesian belief networks: from construction to inference}.
\newblock PhD thesis, 1995.

\bibitem{koivisto2004exact}
Mikko Koivisto and Kismat Sood.
\newblock Exact bayesian structure discovery in bayesian networks.
\newblock {\em The Journal of Machine Learning Research}, 5:549--573, 2004.

\bibitem{ott2003finding}
Sascha Ott, Seiya Imoto, and Satoru Miyano.
\newblock Finding optimal models for small gene networks.
\newblock In {\em Biocomputing 2004}, pages 557--567. World Scientific, 2003.

\bibitem{cussens2012bayesian}
James Cussens.
\newblock Bayesian network learning with cutting planes.
\newblock {\em arXiv preprint arXiv:1202.3713}, 2012.

\bibitem{BARTLETT2017258}
Mark Bartlett and James Cussens.
\newblock Integer linear programming for the bayesian network structure learning problem.
\newblock {\em Artificial Intelligence}, 244:258--271, 2017.
\newblock Combining Constraint Solving with Mining and Learning.

\bibitem{ijcai2017p708}
James Cussens, Matti Järvisalo, Janne~H. Korhonen, and Mark Bartlett.
\newblock Bayesian network structure learning with integer programming: Polytopes, facets and complexity (extended abstract).
\newblock In {\em Proceedings of the Twenty-Sixth International Joint Conference on Artificial Intelligence, {IJCAI-17}}, pages 4990--4994, 2017.

\bibitem{tsamardinos2006max}
Ioannis Tsamardinos, Laura~E Brown, and Constantin~F Aliferis.
\newblock The max-min hill-climbing bayesian network structure learning algorithm.
\newblock {\em Machine learning}, 65:31--78, 2006.

\bibitem{song2022using}
Wenzhu Song, Hao Gong, Qili Wang, Lijuan Zhang, Lixia Qiu, Xueli Hu, Huimin Han, Yaheng Li, Rongshan Li, and Yafeng Li.
\newblock Using bayesian networks with max-min hill-climbing algorithm to detect factors related to multimorbidity.
\newblock {\em Frontiers in Cardiovascular Medicine}, 9:984883, 2022.

\bibitem{constantinou2020learninga}
Anthony~C Constantinou.
\newblock Learning bayesian networks with the saiyan algorithm.
\newblock {\em ACM Transactions on Knowledge Discovery from Data (TKDD)}, 14(4):1--21, 2020.

\bibitem{constantinou2020learningb}
Anthony~C Constantinou.
\newblock Learning bayesian networks that enable full propagation of evidence.
\newblock {\em IEEE Access}, 8:124845--124856, 2020.

\bibitem{johnson1975finding}
Donald~B Johnson.
\newblock Finding all the elementary circuits of a directed graph.
\newblock {\em SIAM Journal on Computing}, 4(1):77--84, 1975.

\bibitem{pearl1988probabilistic}
Judea Pearl.
\newblock {\em Probabilistic reasoning in intelligent systems: networks of plausible inference}.
\newblock Morgan kaufmann, 1988.

\bibitem{scutari2009learning}
Marco Scutari.
\newblock Learning bayesian networks with the bnlearn r package.
\newblock {\em arXiv preprint arXiv:0908.3817}, 2009.

\bibitem{scutari2012bayesian}
Marco Scutari.
\newblock Bayesian network repository.
\newblock {\em URL http://www. bnlearn. com}, 420, 2012.

\bibitem{pmlr-v138-cussens20a}
James Cussens.
\newblock Gobnilp: Learning bayesian network structure with integer programming.
\newblock In Manfred Jaeger and Thomas~Dyhre Nielsen, editors, {\em Proceedings of the 10th International Conference on Probabilistic Graphical Models}, volume 138 of {\em Proceedings of Machine Learning Research}, pages 605--608. PMLR, 23--25 Sep 2020.

\bibitem{neapolitan2004learning}
Richard~E Neapolitan et~al.
\newblock {\em Learning bayesian networks}, volume~38.
\newblock Pearson Prentice Hall Upper Saddle River, 2004.

\end{thebibliography}


\appendix
\onecolumn
\section*{Appendix}

\begin{figure}[h]
\vskip 0.2in
\begin{center}
\centerline{\includegraphics[width=\columnwidth,height=250pt]{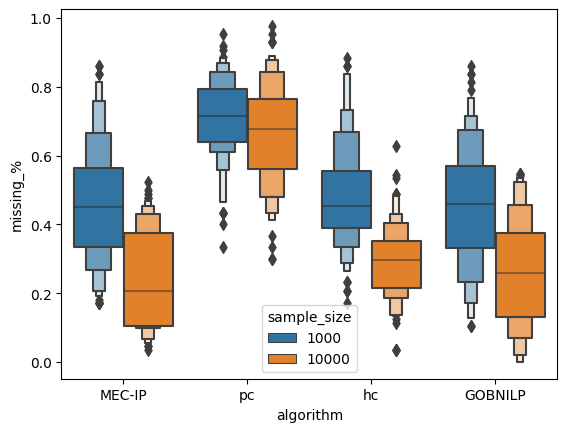}}
\label{missing-bp}
\end{center}
\caption{Missing\_\% Distribution Across Simulations by Sample Size and Algorithm}
\label{missing-size}
\vskip -0.2in
\end{figure}

\begin{figure}[h]
\vskip 0.2in
\begin{center}
\centerline{\includegraphics[width=\columnwidth,height=250pt]{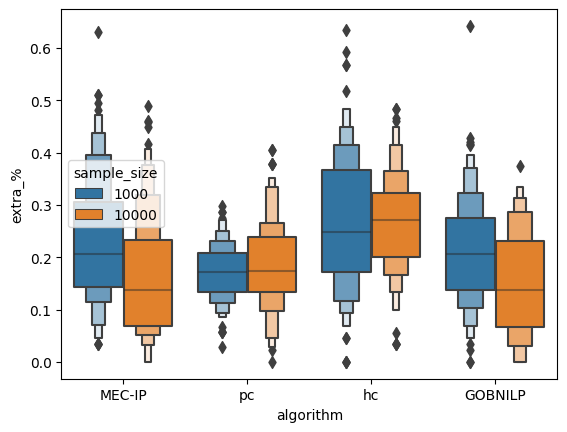}}
\label{missing-bp}
\end{center}
\caption{Extra\_\% Distribution Across Simulations by Sample Size and Algorithm}
\label{extra-size}\vskip -0.2in
\end{figure}

\begin{table}[h]
\centering
\begin{tiny}
\caption{Detailed comparison of network parameters across algorithms and iterations. Values are Mean (std) across 10 replicates.}
\label{Network-table-detailed}
\vskip 0.15in
\begin{tabularx}{0.9\textwidth}{XXXXX}
\toprule
Network, Sample Size & Algorithm & Missing\_\% & Extra\_\% & Tot\_Time \\
\midrule
Alarm, 1000 & MEC-IP & 0.242 (0.029) & \textbf{0.188 (0.067)} & 12.118 (3.898) \\
            & Gobnilp    & \textbf{0.202 (0.044)} & \textbf{0.188 (0.056)} & 4.285 (0.668)  \\
            & HC         & 0.468 (0.029) & 0.496 (0.051) & \textbf{6.164 (0.166)}  \\
            & PC         & 0.376 (0.035) & \textbf{0.188 (0.053)} & 17.686 (1.628) \\
\midrule
Alarm, 10000 & MEC-IP & 0.108 (0.041) & 0.068 (0.047) & 184.237 (75.247) \\
             & Gobnilp    & \textbf{0.082 (0.006)} & \textbf{0.046 (0.013)} & 143.675 (58.195) \\
             & HC         & 0.388 (0.089) & 0.526 (0.087) & \textbf{22.427 (0.432)}  \\
             & PC         & 0.166 (0.033) & 0.094 (0.033) & 59.433 (3.138)  \\
\midrule
Asia, 1000 & MEC-IP & 0.282 (0.151) & \textbf{0.027} (0.044) & 0.215 (0.047) \\
           & Gobnilp    & \textbf{0.273 (0.129)} & 0.036 (0.047) & 0.284 (0.057) \\
           & HC         & 0.482 (0.182) & 0.409 (0.279) & \textbf{0.194 (0.031)} \\
           & PC         & 0.527 (0.112) & 0.255 (0.165) & 0.359 (0.047) \\
\midrule
Asia, 10000 & MEC-IP & \textbf{0.127 (0.088)} & \textbf{0.000 (0.000)} & \textbf{0.499 (0.288)} \\
            & Gobnilp    & \textbf{0.127 (0.088)} & \textbf{0.000 (0.000)} & 1.014 (0.234) \\
            & HC         & 0.373 (0.194) & 0.400 (0.288) & 0.612 (0.056) \\
            & PC         & 0.200 (0.084) & 0.082 (0.100) & 1.383 (0.049) \\
\midrule
Barley, 1000 & MEC-IP & 0.727 (0.026) & 0.270 (0.030) & 6.419 (2.231) \\
             & Gobnilp    & \textbf{0.685 (0.007)} & 0.277 (0.025) & \textbf{2.334 (0.171)} \\
             & HC         & 0.704 (0.026) & 0.338 (0.052) & 8.342 (0.416) \\
             & PC         & 0.698 (0.027) & \textbf{0.198 (0.022)} & 60.713 (7.035) \\
\midrule
Barley, 10000 & MEC-IP & 0.560 (0.050) & 0.168 (0.016) & 341.102 (129.165) \\
              & Gobnilp    & 0.528 (0.006) & 0.180 (0.016) & \textbf{18.534 (5.718)} \\
              & HC         & 0.649 (0.062) & 0.370 (0.051) & 35.437 (1.145) \\
              & PC         & \textbf{0.502 (0.023)} & \textbf{0.120 (0.044)} & 759.509 (92.541) \\
\midrule
Child, 1000 & MEC-IP & 0.194 (0.050) & 0.214 (0.020) & 3.900 (1.815) \\
            & Gobnilp    & \textbf{0.117 (0.065)} & 0.149 (0.071) & \textbf{1.092 (0.232)} \\
            & HC         & 0.371 (0.122) & 0.249 (0.113) & 1.440 (0.064) \\
            & PC         & 0.483 (0.068) & \textbf{0.146 (0.039)} & 17.075 (3.763) \\
\midrule
Child, 10000 & MEC-IP & 0.297 (0.130) & 0.120 (0.050) & 142.757 (73.869) \\
             & Gobnilp    & \textbf{0.040 (0.090)} & \textbf{0.020 (0.033)} & \textbf{4.872 (1.291)} \\
             & HC         & 0.254 (0.143) & 0.180 (0.089) & 5.718 (0.309) \\
             & PC         & 0.409 (0.024) & 0.126 (0.024) & 93.186 (2.445) \\
\midrule
Insurance, 1000 & MEC-IP & \textbf{0.496 (0.048)} & 0.222 (0.080) & 8.082 (3.203) \\
                & Gobnilp    & 0.502 (0.014) & \textbf{0.209 (0.015)} & \textbf{2.340 (0.467)} \\
                & HC         & 0.570 (0.056) & 0.363 (0.068) & 3.057 (0.236) \\
                & PC         & 0.693 (0.033) & 0.243 (0.027) & 20.137 (2.407) \\
\midrule
Insurance, 10000 & MEC-IP & 0.398 (0.102) & 0.185 (0.089) & 384.331 (118.207) \\
                 & Gobnilp    & \textbf{0.248 (0.037)} & \textbf{0.139 (0.029)} & 18.471 (6.636) \\
                 & HC         & 0.500 (0.048) & 0.411 (0.063) & \textbf{13.118 (0.553)} \\
                 & PC         & 0.563 (0.049) & 0.213 (0.065) & 173.999 (18.602) \\
\midrule
Sachs, 1000 & MEC-IP & \textbf{0.316 (0.134)} & \textbf{0.211 (0.089)} & 2.402 (0.971) \\
            & Gobnilp    & 0.332 (0.151) & 0.226 (0.114) & \textbf{0.447 (0.059)} \\
            & HC         & 0.447 (0.122) & 0.353 (0.122) & 0.420 (0.027) \\
            & PC         & 0.495 (0.044) & 0.389 (0.106) & 12.267 (3.656) \\
\midrule
Sachs, 10000 & MEC-IP & \textbf{0.221 (0.095)} & 0.247 (0.117) & 6.476 (0.720) \\
             & Gobnilp    & 0.268 (0.072) & \textbf{0.216 (0.072)} & \textbf{1.023 (0.140)} \\
             & HC         & 0.353 (0.079) & 0.379 (0.092) & 1.701 (0.082) \\
             & PC         & 0.326 (0.022) & 0.284 (0.044) & 33.443 (0.877) \\
\midrule
Water, 1000 & MEC-IP & 0.644 (0.013) & 0.246 (0.030) & 1.829 (0.281) \\
            & Gobnilp    & \textbf{0.618 (0.029)} & 0.218 (0.048) & \textbf{1.299 (0.420)} \\
            & HC         & 0.651 (0.040) & 0.244 (0.044) & 3.259 (0.146) \\
            & PC         & 0.688 (0.019) & \textbf{0.175 (0.015)} & 2.187 (0.161) \\
\midrule
Water, 10000 & MEC-IP & 0.665 (0.048) & 0.200 (0.069) & 309.992 (156.162) \\
             & Gobnilp    & \textbf{0.599 (0.020)} & 0.143 (0.036) & \textbf{13.342 (4.825)} \\
             & HC         & 0.711 (0.039) & 0.353 (0.028) & 13.566 (0.627) \\
             & PC         & 0.638 (0.019) & \textbf{0.132 (0.012)} & 18.503 (1.281) \\
\midrule
Win95pts, 1000 & MEC-IP & \textbf{0.478 (0.023)} & \textbf{0.242 (0.049)} & 137.478 (112.893) \\
               & Gobnilp    & 0.582 (0.031) & 0.508 (0.037) & 328.136 (292.290) \\
               & HC         & 0.576 (0.039) & 0.541 (0.064) & 34.998 (1.083)   \\
               & PC         & 0.634 (0.030) & 0.248 (0.024) & \textbf{23.778 (2.786)}  \\
\midrule
Win95pts, 10000 & MEC-IP & \textbf{0.356 (0.063)} & \textbf{0.189 (0.052)} & 1293.529 (927.761) \\
                & Gobnilp    & 0.514 (0.022) & 0.516 (0.027) & 2387.577 (1786.621) \\
                & HC         & 0.483 (0.070) & 0.668 (0.155) & \textbf{113.818 (15.914)}   \\
                & PC         & 0.439 (0.032) & 0.148 (0.021) & 159.991 (13.913)   \\
\bottomrule
\end{tabularx}
\end{tiny}
\vskip -0.1in
\end{table}

\begin{table}[h]
\caption{Simulation Results for Synthetic 20 - Node BNs}
\centering
\begin{tiny}
\label{20 - Networks}
\vskip 0.15in
\begin{tabularx}{0.9\columnwidth}{XXXXX}
\toprule
Iteration & Algorithm & Missing\_\% & Extra\_\% & Tot\_Time \\
\midrule
(20, 2, 2, 1), 1000 & MEC-IP & 0.207 (0.040) & 0.293 (0.128) & 0.391 (0.082) \\
                    & Gobnilp    & \textbf{0.190 (0.059)} & 0.307 (0.113) & \textbf{0.276 (0.085)} \\
                    & HC         & 0.317 (0.099) & 0.407 (0.051) & 1.287 (0.047) \\
                    & PC         & 0.693 (0.055) & \textbf{0.169 (0.030)} & 1.305 (0.064) \\
\midrule
(20, 2, 2, 1), 10000 & MEC-IP & 0.138 (0.023) & 0.038 (0.047) & \textbf{0.473 (0.081)} \\
                     & Gobnilp    & \textbf{0.045 (0.049)} & \textbf{0.034 (0.049)} & 1.113 (0.261) \\
                     & HC         & 0.338 (0.100) & 0.359 (0.103) & 1.622 (0.075) \\
                     & PC         & 0.431 (0.024) & 0.231 (0.037) & 2.936 (0.258) \\
\midrule
(20, 2, 2, 5), 1000 & MEC-IP & \textbf{0.293 (0.081)} & 0.293 (0.117) & 0.432 (0.098) \\
                    & Gobnilp    & 0.307 (0.073) & 0.227 (0.062) & \textbf{0.146 (0.013)} \\
                    & HC         & 0.380 (0.125) & 0.353 (0.148) & 1.269 (0.059) \\
                    & PC         & 0.663 (0.037) & \textbf{0.130 (0.037)} & 0.671 (0.062) \\
\midrule
(20, 2, 2, 5), 10000 & MEC-IP & 0.117 (0.032) & 0.080 (0.032) & 0.542 (0.126) \\
                     & Gobnilp    & \textbf{0.070 (0.055)} & \textbf{0.027 (0.034)} & \textbf{0.306 (0.034)} \\
                     & HC         & 0.273 (0.100) & 0.290 (0.121) & 1.931 (0.183) \\
                     & PC         & 0.730 (0.058) & 0.077 (0.042) & 1.991 (0.164) \\
\midrule
(20, 2, 4, 1), 1000 & MEC-IP & 0.313 (0.023) & 0.103 (0.055) & 0.308 (0.061) \\
                    & Gobnilp    & \textbf{0.290 (0.022)} & 0.087 (0.061) & \textbf{0.116 (0.017)} \\
                    & HC         & \textbf{0.290 (0.022)} & \textbf{0.083 (0.065)} & 1.011 (0.046) \\
                    & PC         & 0.443 (0.050) & 0.140 (0.031) & 0.947 (0.147) \\
\midrule
(20, 2, 4, 1), 10000 & MEC-IP & \textbf{0.097 (0.011)} & \textbf{0.070 (0.011)} & \textbf{0.503 (0.014)} \\
                     & Gobnilp    & 0.240 (0.014) & 0.073 (0.014) & 0.510 (0.034) \\
                     & HC         & 0.117 (0.061) & 0.167 (0.099) & 1.564 (0.048) \\
                     & PC         & 0.400 (0.068) & 0.150 (0.018) & 5.985 (0.524) \\
\midrule
(20, 2, 4, 5), 1000 & MEC-IP & \textbf{0.417 (0.050)} & 0.245 (0.038) & 0.300 (0.025) \\
                    & Gobnilp    & \textbf{0.417 (0.050)}& 0.245 (0.038) & \textbf{0.065 (0.008)}\\
                    & HC         & 0.428 (0.044) & 0.228 (0.029) & 0.919 (0.028) \\
                    & PC         & 0.631 (0.052) & \textbf{0.166 (0.036)} & 0.608 (0.077) \\
\midrule
(20, 2, 4, 5), 10000 & MEC-IP & \textbf{0.100 (0.011)} & 0.134 (0.011) & 0.549 (0.016) \\
                     & Gobnilp    & \textbf{0.100 (0.011)} & 0.134 (0.011) & \textbf{0.343 (0.069)} \\
                     & HC         & 0.290 (0.080) & 0.328 (0.085) & 1.506 (0.067) \\
                     & PC         & 0.641 (0.071) & \textbf{0.128 (0.037)} & 4.931 (0.870) \\
\midrule
(20, 3, 2, 1), 1000 & Challenger & 0.505 (0.069) & 0.284 (0.104) & \textbf{1.325 (0.262)} \\
                    & Gobnilp    & 0.508 (0.074) & \textbf{0.195 (0.074)} & 4.299 (2.736) \\
                    & HC         & \textbf{0.457 (0.041)} & 0.254 (0.083) & 1.477 (0.064) \\
                    & PC         & 0.841 (0.037) & 0.227 (0.034) & 3.467 (0.797) \\
\midrule
(20, 3, 2, 1), 10000 & Challenger & 0.281 (0.056) & 0.259 (0.081) & 3.716 (0.984) \\
                     & Gobnilp    & 0.408 (0.068) & \textbf{0.181 (0.042)} & 11.493 (5.174) \\
                     & HC         & \textbf{0.276 (0.079)} & 0.324 (0.093) & \textbf{2.320 (0.110)} \\
                     & PC         & 0.716 (0.041) & 0.365 (0.032) & 57.191 (8.868) \\
\midrule
(20, 3, 2, 5), 1000 & Challenger & 0.557 (0.065) & 0.154 (0.063) & 0.217 (0.026) \\
                    & Gobnilp    & 0.577 (0.066) & 0.174 (0.076) & \textbf{0.119 (0.006)} \\
                    & HC         & \textbf{0.571 (0.069)} & 0.169 (0.041) & 0.983 (0.071) \\
                    & PC         & 0.686 (0.019) & \textbf{0.074 (0.024)} & 0.391 (0.033) \\
\midrule
(20, 3, 2, 5), 10000 & Challenger & \textbf{0.340 (0.084)} & 0.146 (0.076) & \textbf{0.610 (0.113)} \\
                     & Gobnilp    & 0.443 (0.041) & 0.260 (0.021) & 0.751 (1.013) \\
                     & HC         & 0.403 (0.067) & 0.240 (0.092) & 1.718 (0.074) \\
                     & PC         & 0.680 (0.040) & \textbf{0.046 (0.031)} & 1.934 (0.640) \\
\midrule
(20, 3, 4, 1), 1000 & MEC-IP & 0.583 (0.063) & 0.179 (0.041) & 0.624 (0.180) \\
                    & Gobnilp    & 0.579 (0.087) & 0.207 (0.056) & \textbf{0.219 (0.051)} \\
                    & HC         & \textbf{0.462 (0.067)} & 0.183 (0.048) & 1.383 (0.112) \\
                    & PC         & 0.788 (0.035) & \textbf{0.171 (0.037)} & 2.149 (0.436) \\
\midrule
(20, 3, 4, 1), 10000 & MEC-IP & 0.433 (0.052) & 0.348 (0.039) & 2.001 (0.670) \\
                     & Gobnilp    & 0.519 (0.031) & 0.314 (0.025) & \textbf{1.691 (0.354)} \\
                     & HC         & \textbf{0.267 (0.082)} & \textbf{0.224 (0.067)} & 2.297 (0.129) \\
                     & PC         & 0.810 (0.039) & 0.257 (0.035) & 30.812 (5.640) \\
\midrule
(20, 3, 4, 5), 1000 & MEC-IP & 0.821 (0.027) & 0.077 (0.027) & 0.239 (0.022) \\
                    & Gobnilp    & \textbf{0.791 (0.045)} & \textbf{0.070 (0.033)} & \textbf{0.068 (0.005)} \\
                    & HC         & 0.828 (0.037) & 0.079 (0.025) & 0.755 (0.028) \\
                    & PC         & 0.870 (0.052) & 0.133 (0.033) & 0.548 (0.077) \\
\midrule
(20, 3, 4, 5), 10000 & MEC-IP & \textbf{0.428 (0.011)} & \textbf{0.100 (0.011)} & \textbf{0.839 (0.016)} \\
                     & Gobnilp    & 0.451 (0.011) & 0.160 (0.011) & 1.520 (0.069) \\
                     & HC         & 0.444 (0.080) & 0.244 (0.085) & 2.006 (0.067) \\
                     & PC         & 0.909 (0.071) & 0.086 (0.037) & 10.051 (0.870) \\
\bottomrule
\end{tabularx}
\end{tiny}
\vskip -0.1in
\end{table}

\begin{table}[h]
\centering
\caption{Simulation Results for Synthetic 60 - Node BNs}
\begin{tiny}
\label{60 - Networks}
\vskip 0.15in
\begin{tabularx}{0.9\textwidth}{XXXXX}
\toprule
Iteration & Algorithm & Missing\_\% & Extra\_\% & Tot\_Time \\
\midrule
(60, 2, 2, 1), 1000 & MEC-IP & 0.355 (0.078) & 0.299 (0.120) & 3.287 (1.034) \\
                    & Gobnilp    & \textbf{0.331 (0.036)}& \textbf{0.231 (0.072)}& \textbf{3.228 (2.184)} \\
                    & HC         & 0.428 (0.071) & 0.385 (0.063) & 19.035 (6.980) \\
                    & PC         & 0.605 (0.029) & 0.256 (0.032) & 7.962 (1.989) \\
\midrule
(60, 2, 2, 1), 10000 & MEC-IP & 0.214 (0.058) & 0.177 (0.114) & \textbf{5.121 (1.708)} \\
                     & Gobnilp    & \textbf{0.136 (0.036)} & \textbf{0.106 (0.043)} & 37.128 (13.808) \\
                     & HC         & 0.330 (0.050) & 0.347 (0.052) & 26.897 (9.965) \\
                     & PC         & 0.502 (0.030) & 0.266 (0.034) & 75.548 (21.849) \\
\midrule
(60, 2, 2, 5), 1000 & MEC-IP & 0.491 (0.069) & 0.452 (0.078) & 2.663 (0.916) \\
                    & Gobnilp    & \textbf{0.485 (0.048)} & 0.380 (0.102) & \textbf{1.547 (0.477)} \\
                    & HC         & 0.488 (0.057) & 0.442 (0.098) & 17.884 (6.165) \\
                    & PC         & 0.736 (0.024) & \textbf{0.194 (0.019)} & 4.220 (1.107) \\
\midrule
(60, 2, 2, 5), 10000 & MEC-IP & 0.249 (0.089) & 0.228 (0.113) & \textbf{4.393 (1.503)} \\
                     & Gobnilp    & \textbf{0.170 (0.032)} & \textbf{0.132 (0.048)} & 5.174 (1.937) \\
                     & HC         & 0.288 (0.035) & 0.268 (0.056) & 25.784 (9.774) \\
                     & PC         & 0.599 (0.026) & 0.177 (0.018) & 10.370 (3.236) \\
\midrule
(60, 2, 4, 1), 1000 & MEC-IP & 0.238 (0.040) & 0.169 (0.038) & 3.120 (0.663) \\
                    & Gobnilp    & \textbf{0.163 (0.041)} & \textbf{0.132 (0.036)} &\textbf{1.323 (0.092)} \\
                    & HC         & 0.368 (0.038) & 0.263 (0.033) & 14.279 (0.680) \\
                    & PC         & 0.729 (0.028) & 0.238 (0.012) & 11.756 (1.410) \\
\midrule
(60, 2, 4, 1), 10000 & MEC-IP & 0.103 (0.017) & 0.141 (0.040) & 48.842 (20.701) \\
                     & Gobnilp    & \textbf{0.046 (0.055)} & \textbf{0.081 (0.056)} & \textbf{13.694 (5.183)} \\
                     & HC         & 0.236 (0.024) & 0.201 (0.035) & 20.015 (0.939) \\
                     & PC         & 0.641 (0.025) & 0.248 (0.021) & 83.739 (5.294) \\
\midrule
(60, 2, 4, 5), 1000 & MEC-IP & \textbf{0.378 (0.041)} & 0.152 (0.022) & 1.686 (0.110) \\
                    & Gobnilp    & 0.427 (0.076) & 0.158 (0.025) & \textbf{0.981 (0.100)} \\
                    & HC         & 0.446 (0.056) & 0.144 (0.027) & 9.350 (0.373) \\
                    & PC         & 0.664 (0.026) & \textbf{0.122 (0.011)} & 3.719 (0.194) \\
\midrule
(60, 2, 4, 5), 10000 & MEC-IP & \textbf{0.085 (0.088)} & 0.049 (0.053) & \textbf{2.493 (0.385)} \\
                     & Gobnilp    & 0.204 (0.029) & \textbf{0.025 (0.015)} & 2.994 (0.109) \\
                     & HC         & 0.184 (0.047) & 0.145 (0.052) & 16.411 (0.649) \\
                     & PC         & 0.563 (0.031) & 0.143 (0.008) & 10.805 (0.316) \\
\midrule
(60, 3, 2, 1), 1000 & MEC-IP & 0.439 (0.056) & 0.248 (0.044) & 2.639 (0.459) \\
                    & Gobnilp    & 0.455 (0.029) & 0.296 (0.036) & \textbf{2.069 (0.549)} \\
                    & HC         & \textbf{0.415 (0.043)} & 0.329 (0.050) & 14.634 (0.529) \\
                    & PC         & 0.637 (0.020) & \textbf{0.187 (0.018)} & 4.149 (0.213) \\
\midrule
(60, 3, 2, 1), 10000 & MEC-IP & 0.267 (0.069) & 0.220 (0.086) & 180.383 (236.930) \\
                     & Gobnilp    & 0.348 (0.007) & 0.249 (0.011) & 31.884 (16.343) \\
                     & HC         & \textbf{0.264 (0.051)} & 0.283 (0.042) & 20.956 (1.145) \\
                     & PC         & 0.581 (0.037) & \textbf{0.193 (0.017)} & \textbf{16.059 (0.986)} \\
\midrule
(60, 3, 2, 5), 1000 & MEC-IP & 0.599 (0.052) & 0.358 (0.107) & 1.910 (0.132) \\
                    & Gobnilp    & \textbf{0.560 (0.043)} & 0.314 (0.072) & \textbf{1.327 (0.228)} \\
                    & HC         & 0.583 (0.052) & 0.392 (0.070) & 12.119 (0.561) \\
                    & PC         & 0.812 (0.027) & \textbf{0.211 (0.020)} & 3.392 (0.266) \\
\midrule
(60, 3, 2, 5), 10000 & MEC-IP & 0.392 (0.081) & 0.331 (0.131) & 136.043 (140.459) \\
                     & Gobnilp    & 0.366 (0.019) & 0.319 (0.025) & 21.135 (11.511) \\
                     & HC         & \textbf{0.246 (0.042)} & 0.300 (0.043) & 21.303 (0.832) \\
                     & PC         & 0.789 (0.022) & \textbf{0.220 (0.021)} & \textbf{11.352 (1.317)} \\
\midrule
(60, 3, 4, 1), 1000 & MEC-IP & \textbf{0.448 (0.031)} & 0.207 (0.074) & 2.522 (0.360) \\
                    & Gobnilp    & 0.502 (0.046) & 0.182 (0.038) & \textbf{1.016 (0.064)} \\
                    & HC         & 0.524 (0.036) & 0.316 (0.054) & 12.278 (0.578) \\
                    & PC         & 0.745 (0.017) & \textbf{0.172 (0.014)} & 7.416 (0.989) \\
\midrule
(60, 3, 4, 1), 10000 & MEC-IP & \textbf{0.228 (0.098)} & \textbf{0.139 (0.106)} & 242.869 (513.812) \\
                     & Gobnilp    & 0.267 (0.018) & 0.189 (0.014) & \textbf{6.633 (1.918)} \\
                     & HC         & 0.299 (0.045) & 0.290 (0.049) & 20.677 (0.642) \\
                     & PC         & 0.720 (0.022) & 0.162 (0.018) & 48.690 (7.089) \\
\midrule
(60, 3, 4, 5), 1000 & MEC-IP & 0.693 (0.031) & 0.191 (0.064) & 1.919 (0.229) \\
                    & Gobnilp    & \textbf{0.682 (0.025)} & 0.197 (0.053) & \textbf{1.044 (0.129)} \\
                    & HC         & 0.706 (0.017) & 0.221 (0.036) & 9.105 (0.599) \\
                    & PC         & 0.864 (0.015) & \textbf{0.158 (0.031)} & 4.312 (0.319) \\
\midrule
(60, 3, 4, 5), 10000 & MEC-IP & 0.424 (0.030) & 0.227 (0.048) & 142.870 (168.419) \\
                     & Gobnilp    & \textbf{0.374 (0.015)} & \textbf{0.114 (0.023)} & \textbf{3.811 (0.325)} \\
                     & HC         & 0.382 (0.032) & 0.238 (0.046) & 19.885 (1.077) \\
                     & PC         & 0.868 (0.014) & 0.167 (0.018) & 96.092 (18.335) \\
\bottomrule
\end{tabularx}
\end{tiny}
\vskip -0.1in
\end{table}

\bibliographystyle{unsrt}  


\end{document}